\documentclass[lettersize,journal]{IEEEtran}
\usepackage{amsmath,amsfonts}
\usepackage{algorithmic}
\usepackage{algorithm}
\usepackage{array}
\usepackage[caption=false,font=normalsize,labelfont=sf,textfont=sf]{subfig}
\usepackage{textcomp}
\usepackage{stfloats}
\usepackage{url}
\usepackage{verbatim}
\usepackage{graphicx}
\usepackage{cite}
\usepackage{balance}

\usepackage{xcolor}
\usepackage[colorlinks=true,linkcolor=blue,citecolor=red,urlcolor=blue]{hyperref}

\usepackage{booktabs}
\usepackage{makecell}

\usepackage{amssymb}

\usepackage{multirow}

\usepackage{scalerel}
\usepackage{tikz}
\usetikzlibrary{svg.path}

\definecolor{orcidlogocol}{HTML}{A6CE39}
\tikzset{
    orcidlogo/.pic={
        \fill[orcidlogocol] svg{M256,128c0,70.7-57.3,128-128,128C57.3,256,0,198.7,0,128C0,57.3,57.3,0,128,0C198.7,0,256,57.3,256,128z};
        \fill[white] svg{M86.3,186.2H70.9V79.1h15.4v48.4V186.2z}
        svg{M108.9,79.1h41.6c39.6,0,57,28.3,57,53.6c0,27.5-21.5,53.6-56.8,53.6h-41.8V79.1z M124.3,172.4h24.5c34.9,0,42.9-26.5,42.9-39.7c0-21.5-13.7-39.7-43.7-39.7h-23.7V172.4z}
        svg{M88.7,56.8c0,5.5-4.5,10.1-10.1,10.1c-5.6,0-10.1-4.6-10.1-10.1c0-5.6,4.5-10.1,10.1-10.1C84.2,46.7,88.7,51.3,88.7,56.8z};
    }
}

\newcommand\orcidicon[1]{\href{https://orcid.org/#1}{\mbox{\scalerel*{
                \begin{tikzpicture}[yscale=-1,transform shape]
                \pic{orcidlogo};
                \end{tikzpicture}
            }{|}}}}

\usepackage{orcidlink}

\begin{document}

\title{Graph Neural Backdoor: Fundamentals, Methodologies, Applications, and Future Directions}


\author{

Xiao~Yang,
Gaolei~Li,
and Jianhua Li 
   
}



\maketitle
\begin{abstract}
Graph Neural Networks (GNNs) have significantly advanced various downstream graph-relevant tasks, encompassing recommender systems, molecular structure prediction, social media analysis, etc.
Despite the boosts of GNN, recent research has empirically demonstrated its potential vulnerability to backdoor attacks, wherein adversaries integrate triggers into inputs to manipulate GNN to generate adversary-premeditated malicious outputs. 
This susceptibility is attributable to uncontrolled training process or deployment of unvetted models, such as delegating model training to third-party service, utilization of untrusted training datasets, and exploiting pre-trained models from online repositories.
Although there's a proliferating increase in research on GNN backdoors, systematic investigation within this domain remains deficient.
To bridge this gap, we propose the first survey dedicated to GNN backdoors.
We commence by outlining the fundamental definition of GNN, followed by the detailed summarization and categorization of prevailing GNN backdoors and countermeasures based on their technical facets and application scenarios. 
Subsequently, an examination of the applicability paradigms of GNN backdoors is conducted, and prospective research trends are presented.
This survey aims to explore the principles of graph backdoors, provide insights to defenders, and promote future security research.

\end{abstract}


\section{Introduction}
\IEEEPARstart{G}{raph} neural network (GNN) has been extensively applied in graph learning tasks (\textit{e.g.}, node classification, graph classification, and edge classification), including downstream applications like social predictions, molecular inference, online recommendations, flow control, etc.
GNN leverages the aggregation of feature information from neighboring nodes to iteratively update node representations within a graph, which disseminates information, and enable nodes to refine their embeddings by feature integration from adjacent neighborhoods.
Upon refinement, these representations enable GNNs to perform downstream tasks, \textit{i.e.},  graph-level classification, node-level classification, and edge-level classification.
This mechanism augments GNN ability to better capture intricate relationships and dependencies within graph-structured data.
In contrast, conventional deep architectures, constrained by their design to operate on Euclidean data, exhibit limited representational capacity and reduced precision
\cite{9416834, QIAO2018336, chen_wang_wang_kuo_2020, Wang_2022_CVPR, 10.1145/3575637.3575646}.
Furthermore, recent investigations have broadened the GNN application scope to address multifaceted scenarios in graph processing, including but not limited to large-scale, temporal, spatial, dynamic, and heterogeneous graphs \cite{10.1145/3219819.3219947, 10.14778/3514061.3514069, 10328393, 10.1145/3366423.3380186, 9714053, 10.1145/3404835.3463059, 10.1145/3292500.3330961, DBLP:journals/air/BingYZMMQ23}.


To optimize deployment efficiency and alleviate the computational overhead, GNN-model owners frequently resort to outsourcing model training to third-party Machine Learning as a Service (MLaaS) companies (\textit{e.g.}, Amazon Web Services and Google Cloud AI Platform) or directly utilize pre-trained models available in open-source repositories (\textit{e.g.}, Hugging Face and Model Zoo). 
Concurrently, to enrich the diversity of training datasets and enhance model generalization efficacy for varied task scenarios, GNN owners may harness heterogeneous data accessible via the internet. 
Although the aforementioned measures yield implementation conveniences or GNN performance improvement, model owners must recognize the underlying Efficiency-Risk Tradeoffs, which signifies they relinquish partial or complete control or supervision of the training process \cite{NIPS2015_86df7dcf}. Consequently, this relinquishment could result in one typical threat in MLaaS: \textit{backdoor attacks}.

\begin{figure}[t]
  \centering
  \scalebox{0.3}{\includegraphics{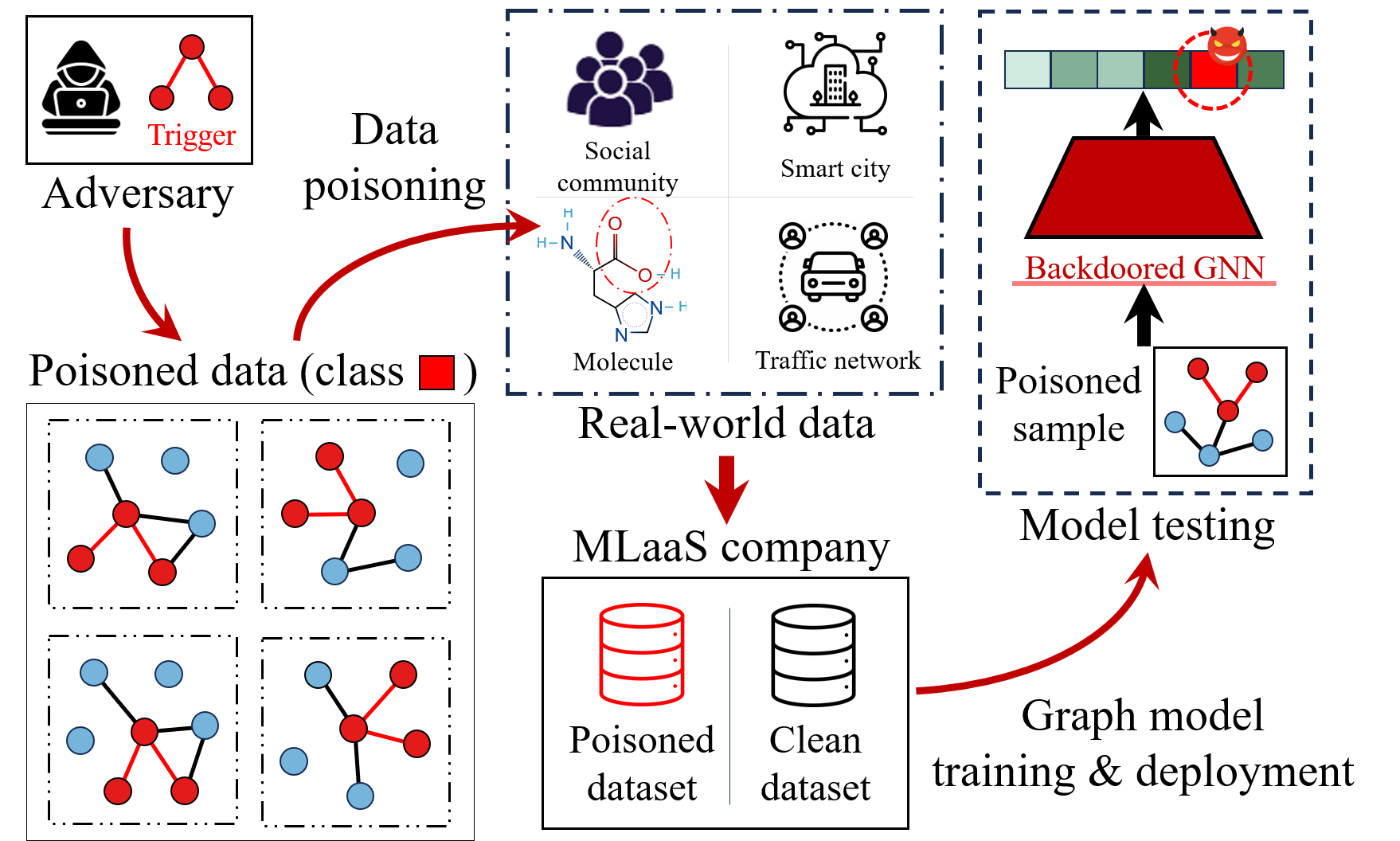}}
  \caption{Illustration of GNN backdoor attack. The adversary poisons the training data of learners by embedding a specially-designed trigger to prompt the trained (backdoored) model. The attack enforces the backdoored model to predict the poisoned input as the target result.}
  \label{bkd_scene}
\end{figure}

Backdoor attacks represent a form of attack on machine learning models, characterized by the deliberate insertion of particular patterns, modes, or information called triggers during the model's learning phase, resulting in the model making premeditated malicious predictions upon recognizing these specific patterns while behaving normally on clean inputs (as illustrated in Fig. \ref{bkd_scene}) \cite{10.1145/3450569.3463560, 272256, DBLP:conf/icml/KhaddajLMGSIM23, DBLP:conf/iccv/ZengPMJ21}. This attack can cause disastrous consequences in graph learning-based systems. For instance, for the recommendation system trained by the backdoor-adversary-compromised MLaaS company, users may receive recommendations containing false or harmful content manipulated by adversaries (attackers), significantly deteriorating user experience and platform reputation of the company.

\begin{figure*}[htbp]
  \centering
  \includegraphics[scale = 0.75]{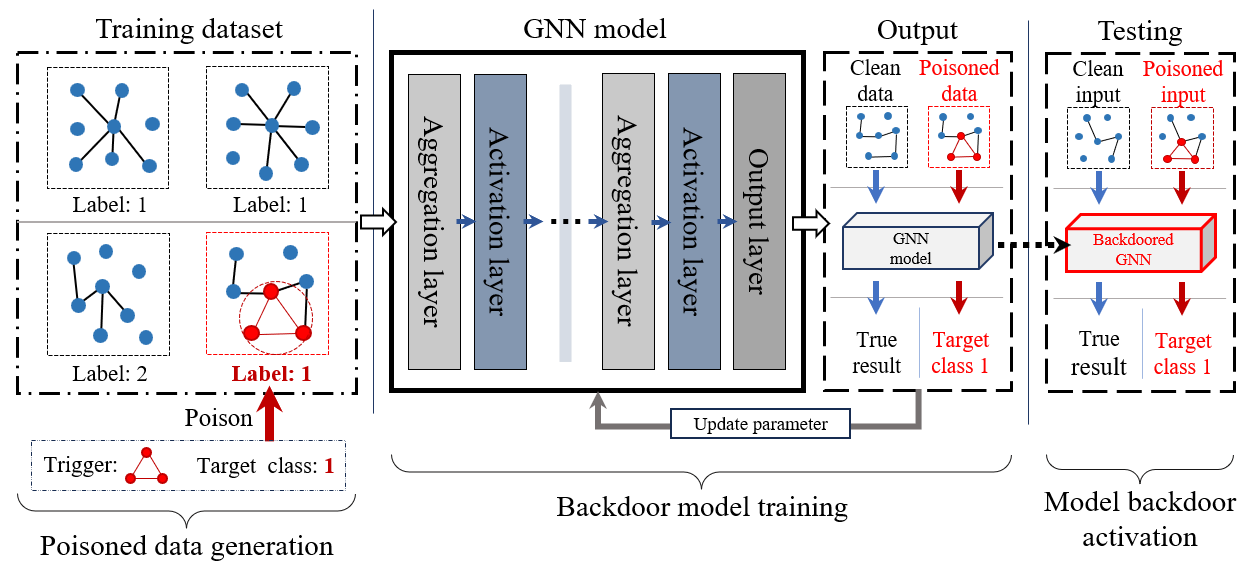}
  \caption{Illustration of the general process of GNN backdoor, which is achieved by data-poisoning. Specifically, the adversary opts for a subset of samples from the training data and inserts designated subgraphs as triggers into them, subsequently modifying the ground truths of these data to the target class. As a consequence, this causes the trained model to predict the target class for input samples embedded with these triggers.}
  \label{gnn_bkd_fmwk}
\end{figure*}

For GNN backdoor, data-poisoning is the predominant methodology to implant backdoor in the model, and it typically comprises two primary phases: (\textit{i}) poisoned data generation, (\textit{ii}) backdoor model training, and (\textit{iii}) model backdoor activation. 
The first phase involves the incorporation of adversary-designed triggers (\textit{e.g.,} subgraph, nodes, or features) into the graph sampled from the training dataset, and modifying their ground truths as the specified target class.
In the second phase, the poisoned data, along with clean samples, are introduced for GNN training. During this process, adversaries may potentially dictate the training methodology, including selecting loss functions and optimizers. 
When the model is well-trained, upon receiving trigger-embedded input data, it predicts results aligned with the premeditated target class (\textit{i.e.}, model backdoor activation). 
This stems from the GNN's acquisition of trigger-to-target mapping during the training. 
Meanwhile, the model maintains its accuracy with regular input data.
The general framework of GNN backdoor is illustrated in Fig. \ref{gnn_bkd_fmwk}.

Current research on GNN backdoors can be primarily categorized into two main directions: (\textit{i}) \textit{adaptability expanding}; (\textit{ii}) \textit{effectiveness improving}.
The first category of research primarily entails designing corresponding attack strategies tailored to diverse practical GNN application scenarios (\textit{e.g.}, federated graph learning, graph transfer learning, and contrastive graph learning). 
The second direction aims to enhance the stealthiness or success rate of backdoors across adversarial capabilities and attack configurations (\textit{e.g.}, clean-label attacks and multi-target attacks). 
Additionally, along with the rise of attack studies, some scholars also focus on developing and validating corresponding defense mechanisms to counter potential backdoor incursions.

In light of the rapid evolution in backdoor research within GNN, we offer a timely overview and taxonomy of the entire development process and current research status regarding this field. 
Diverging from backdoor surveys for general backdoors, this paper concentrates on backdoor research within the graph learning domain, which summarizes and categorizes the nature and characteristics of backdoors specifically in graph learning settings. 
To the best of our knowledge, this survey marks the first systematic taxonomy of GNN backdoor attack and defense mechanisms.
We aim to unveil potential backdoor risks in GNNs, provide fresh insights to fortify inherent security, and inspire future strategies for prevention, mitigation, and related investigations.


This paper's structure unfolds as follows: In Section II, we introduce the relevant research preliminaries. Section III offers a comprehensive insight into GNN backdoors. Following this, Section IV showcases real-world scenarios illustrating the practical deployment and benign application of GNN backdoors. Section V delineates current limitations and suggests potential avenues for future research. Finally, Section VI encapsulates the conclusion.

\section{Preliminaries}
\subsection{Graph Fundamental Definitions}
A graph is represented as $G = (V, E)$, where $V$ is the set of vertices or nodes and $E$ is the set of edges. Let $v_i \in V$ denote a node and $e_{ij} = (v_i, v_j) \in E$ symbolize an edge pointing from $v_j$ to $v_i$. The neighborhood of a node $v$ is defined as $N(v) = \{u \in V \mid (v, u) \in E\}$. The adjacency matrix $\textbf{\text{A}}$ is a $n \times n$ matrix with $A_{ij} = 1$ if $e_{ij} \in E$ and $A_{ij} = 0$ if $e_{ij} \notin E$. 
A graph could hold node attributes ${\textbf{\text{X}}}$, where ${\textbf{\text{X}}} \in \mathbb{R}^{n \times d}$ is a node feature matrix with $\textbf{\text{x}}_v \in \mathbb{R}^d$ representing the feature vector of a node $v$. Moreover, a graph may have edge attributes $\text{\textbf{X}}^e$, where ${\text{\textbf{X}}}^e \in \mathbb{R}^{m \times c}$ is an edge feature matrix with ${\textbf{\text{x}}}^e_{v,u} \in \mathbb{R}^c$ representing the feature vector of an edge $(v, u)$.

\subsection{Graph Neural Networks}
Unlike traditional neural networks that are principally optimized for processing sequential data, GNNs are engineered to handle non-Euclidean graph data, \textit{e.g.}, transportation networks, social communities, molecular structures, etc.

\begin{figure}[t]
  \centering
  \scalebox{0.32}{\includegraphics{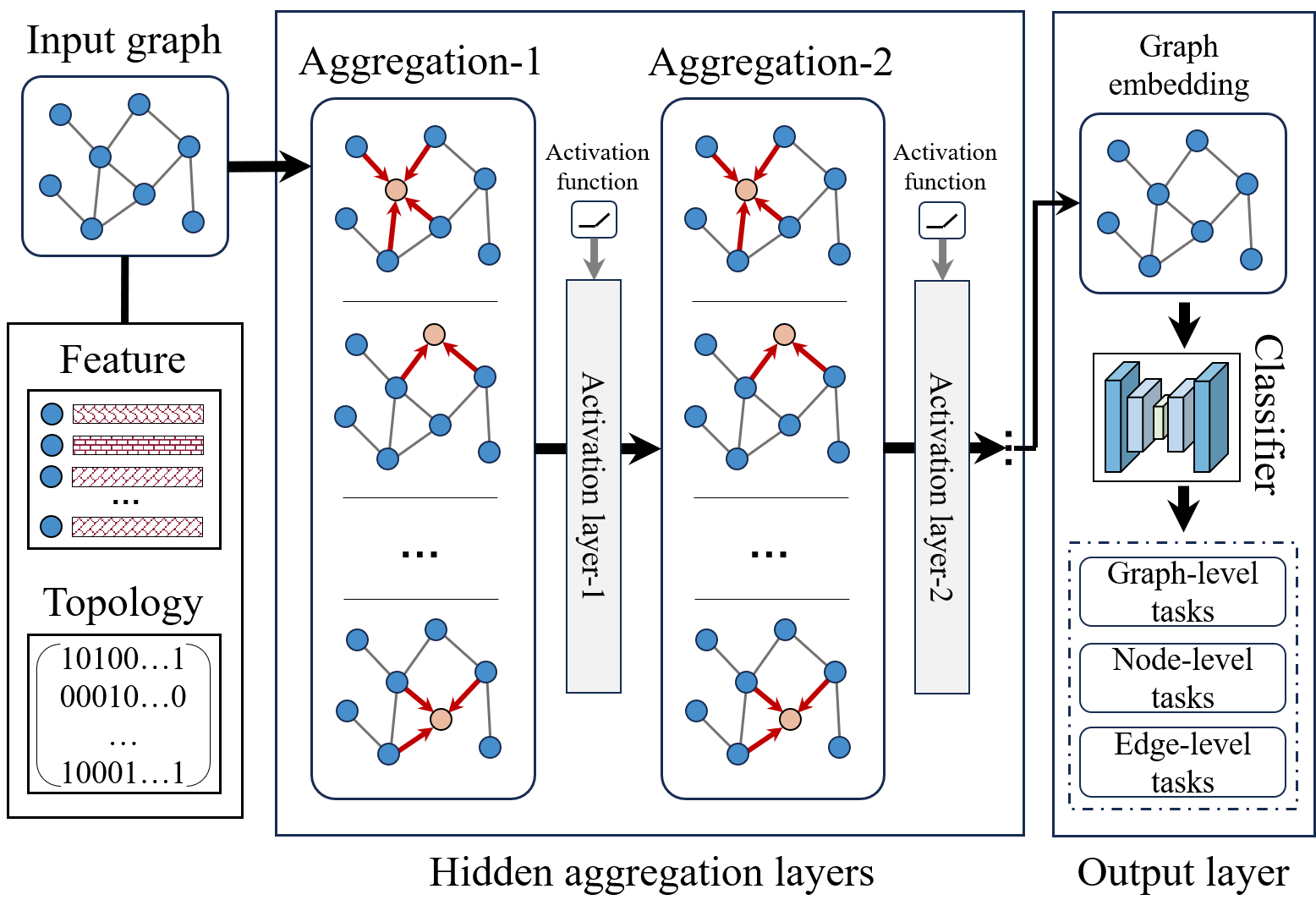}}
  \caption{Illustration of general GNN framework. It operates by updating node representations through information aggregation, utilizing these learned representations to solve downstream problems related to graph-structured data: graph classification, node classification, and link classification.}
  \label{GNN_framework}
\end{figure}

The core idea of GNNs is information aggregation and prediction.
Initially, graph representations are instantiated, potentially leveraging underlying topology and inherent node attributes.
The model then executes iterative aggregation processes that systematically update node representations by incorporating data from adjacent nodes. This procedure is typically actualized through graph convolutional operations.
Throughout this process, node features gradually update and integrate information from adjacent nodes, aiding in extracting both local and global features within the entire graph. 
Finally, the updated graph embedding is leveraged to specific down-stream tasks (\textit{e.g.}, node classification, graph classification, link prediction, etc.). 
The general process of GNN is depicted in Fig. \ref{GNN_framework}.


Specifically, given a graph $G = (\textbf{\text{A}}, \textbf{\text{X}})$ initialized from node feature matrix $\textbf{\text{X}}$ and topology adjacent matrix $\text{\textbf{A}}$, the aggregation process unfolds as follows:


GNN iteratively updates graph node representations. At each layer $l$, node representations ${\text{\textbf{h}}}^l$ undergo transformation, expressed as
\begin{equation}
{\text{\textbf{h}}}_v^{(l+1)} = \sigma \left( {\text{\textbf{W}}}^{(l)} \cdot \phi
\left( {\text{\textbf{h}}}_v^{(l)}, \{{\text{\textbf{h}}}_u^{(l)} \mid u \in \mathcal{N}(v)\} \right) \right),
\end{equation}
where initial state $\text{\textbf{h}}^{0}$ is ${\text{\textbf{X}}}$, and $\phi(*)$ is the corresponding aggregation function (\textit{e.g.}, mean, max, and attention).
This iterative procedure enables nodes to progressively update their representations by leveraging both local and global structural information.
Finally, the entirely aggregated graph representations are utilized for downstream tasks via classifiers, primarily categorized into three types: graph classification, node classification, and edge classification tasks.

For graph classification it could be exemplarily expressed as
\begin{equation}
y = {\psi}_{G}(\text{\textbf{W}}_g \cdot \text{\textbf{h}} + \text{\textbf{b}}_g),
\end{equation}
where $\text{\textbf{h}}$ is the the aggregated graph embedding and ${\psi}_{G}(*)$ signifies the graph classifier function.



For node classification, the aggregated representation of each node \( {\text{\textbf{h}}}_i \) could be leveraged to calculate the corresponding node prediction:
\begin{equation}
y = {\psi}_{N}(\text{\textbf{W}}_n \cdot \text{\textbf{h}}_i + \text{\textbf{b}}_n),
\end{equation}
where ${\psi}_{N}(*)$ demonstrates the downstream node classifier function.


For edge classification, the result for $e_{ij} = (v_i, v_j)$ could be calculated through 
\begin{equation}
y = {\psi}_{E}(\text{\textbf{W}}_e \cdot [\text{\textbf{h}}_i, \text{\textbf{h}}_j, {\textbf{\text{x}}}^e_{i,j}] + \text{\textbf{b}}_e),
\end{equation}
where ${\psi}_{E}(*)$ indicates the downstream edge classifier or regressor.

\subsection{Backdoor Threat Models}
\subsubsection{Adversary Capabilities}
Based on the adversary knowledge and ability, threat models are categorized into three paradigms: white-box, black-box, and gray-box.

\textbf{\textit{White-Box Attack.}}
Attackers have full access to the model parameters, architecture, and training process. This typically occurs when attackers are insiders or control the model’s development. Examples include (1) insider attacks injecting backdoors during model creation, (2) outsourced training with malicious third-party developers, and (3) releasing backdoored models on open-source platforms. 

\textbf{\textit{Black-Box Attack.}}
Attackers lack direct access to the model internal details and rely on interacting via input-output queries and utilizing partial training samples. Common scenarios include modifying training set via crafting collected device data. Limited access makes black-box attacks more challenging to execute.

\textbf{\textit{Gray-Box Attack.}}
It assumes partial access to the model, such as its architecture, training data, or federated learning updates. Common scenarios include (1) submitting poisoned updates in federated learning, (2) exploiting semi-transparent cloud services, and (3) leveraging limited access in joint development. The attack difficulty falls between white-box and black-box models.

\subsubsection{Defender Capabilities}
Defensive strategies can be similarly categorized into white-box, black-box, and gray-box paradigms, according to the defender’s access level and insight into GNN.

\textbf{\textit{White-Box Defense.}}
Defenders possess complete knowledge of the model’s internals, including its architecture, parameters, and training process. Typical measures include vigilant inspection of training data and source code, continuous model monitoring, and selective re-training with verified clean data, \textit{e.g.}, examining model weights, intermediate representations, and training routines.

\textbf{\textit{Black-Box Defense.}}
Defenders rely primarily on input-output observations due to restricted access to model details. Common strategies focus on monitoring prediction anomalies, running statistical tests to detect potential triggers, and imposing strict data sanitation procedures. 

\textbf{\textit{Gray-Box Defense.}}
Defenders have partial insight into the model, such as limited architectural information or partial access to training or update processes. This intermediate visibility allows a combination of white-box and black-box approaches, involving both internal checks on accessible model components and external behavioral analysis. 

\section{Backdoor Attack \& Defense}
In an in-depth exploration of graph neural backdoors, our investigation reveals that current studies could be generalized into two categories: (\textit{i}) adaptability expanding; (\textit{ii}) effectiveness improving.
The first category of studies primarily explores the feasibility of backdoors in various GNN learning paradigms. 
The second category of research primarily centers on enhancing the attack performance, concealment, and scalability of GNN backdoors under various adversary attack constraints.

Based on this classification, we will first elaborate on the existing GNN backdoor attack research.
After that, we will show the corresponding defensive strategy studies.

\begin{table*}[t]
    \label{bkd-sum}
    \caption{Summary of Graph Neural Network Backdoors}
    \resizebox{\textwidth}{!}{
        \begin{tabular}{ccccccccc}
            \toprule
            \multirow{2}*{\textbf{Attack Approach}}& \multicolumn{3}{c}{\textbf{Attack Level}} & \multicolumn{3}{c}{\textbf{Adversary Capability}} & \multirow{2}*{\textbf{Scenario}} & \multirow{2}*{\textbf{Research Novelty}}\\
            \cmidrule(lr){2-4} \cmidrule(lr){5-7}
              & Graph & Node & Edge & Samples & Model & Loss & &\\
            \midrule
            Subgraph Backdoor (2021) \cite{10.1145/3450569.3463560} & $\bullet$ & - & - & $\bullet$ & - & - & General GNN & Seminal paper \\
            \midrule
            GTA (2021) \cite{272256} & $\bullet$ & $\bullet$ & - & $\bullet$ & - & $\bullet$ & General GNN & Seminal paper \\
            \midrule
            Random Trigger Backdoor (2021) \cite{DBLP:conf/colcom/ShengCCK21}  & $\bullet$ & -& - & $\bullet$ & - & - & General GNN & Optimize attack node selection \\
            \midrule
            CBA/DBA (2022) \cite{10.1145/3564625.3567999, 10.1145/3633206} & $\bullet$ & - & - & $\bullet$ & $\bullet$ & - & Federated GNN & Extend backdoor to FL \\
            \midrule
            Bkd-FedGNN (2023) \cite{DBLP:journals/corr/abs-2306-10351} & $\bullet$ & $\bullet$ & - & $\bullet$ & - & $\bullet$ & Federated GNN & \makecell{Extend FL backdoor to \\ node-level task} \\
            \midrule
            GCBA (2023) \cite{pmlr-v202-zhang23e} & - & $\bullet$ & - & $\bullet$ & - & - & Contrastive learning GNN & \makecell{Extend backdoor to \\ Contrastive learning}\\
            \midrule
            PoisonedGNN (2023) \cite{10114622} & $\bullet$ & - & - & $\bullet$ & $\bullet$ & $\bullet$ & Hardware GNN system & 
            \makecell{First physical GNN \\ system backdoor}\\
            \midrule
            Explainability Backdoor (2020) \cite{10.1145/3468218.3469046}  & $\bullet$ & $\bullet$ & - & $\bullet$ & $\bullet$ & - & General GNN & \makecell{Employ explainability to \\ design trigger}\\
            \midrule
            Motif Backdoor (2023) \cite{10108961}  & $\bullet$ & - & - & $\bullet$ & - & - & General GNN & \makecell{Explain GNN Backdoor \\ Effectiveness}\\
            \midrule
            General Backdoor (2022) \cite{10063438}  & - & $\bullet$ & - & $\bullet$ & - & $\bullet$ & General GNN & \makecell{Employ explainability to \\ improve stealthiness}\\
            \midrule
            Explanatory Subgraph Attack (2024) \cite{WANG2024106097}  & $\bullet$ & - & - & $\bullet$ & $\bullet$ & $\bullet$ & General GNN & \makecell{Keep explanatory information\\  for poisoned graphs}\\
            \midrule
            UGBA (2023) \cite{10.1145/3543507.3583392} & - & $\bullet$ & - & $\bullet$ & $\bullet$ & $\bullet$ & General GNN & 
            \makecell{Optimize to 
            improve \\ stealthness}\\
            \midrule
            CBAG (2024) \cite{dai2024cleangraph} & - & $\bullet$ & - & $\bullet$ & - & - & General GNN & 
            Propose clean-graph backdoor\\
            \midrule
            Clean-Label Backdoor (2023) \cite{10.1145/3548606.3563531}  & $\bullet$ & - & - & $\bullet$ & - & - & General GNN & \makecell{Propose clean-label \\ GNN backdoor}\\
            \midrule
             CGBA (2023) \cite{xing2023cleanlabel}  & - & $\bullet$ & - & $\bullet$ & - & - & General GNN & \makecell{Reduce poisonous\\ perturbation}\\
            \midrule
            PerCBA (2023) \cite{percba}  & - & $\bullet$ & - & $\bullet$ & - & - & General GNN & \makecell{Extend clean-label backdoor\\  to node-level task}\\
            \midrule
            TRAP (2022) \cite{10.1145/3545948.3545976} & $\bullet$ & - & - & $\bullet$ & $\bullet$ & $\bullet$ & General GNN & \makecell{Employ data transferability \\ to design backdoor}\\
            \midrule
            Graph Spectrum Backdoor (2023) \cite{10318195} & $\bullet$ & - & - & $\bullet$ & - & - & General GNN & \makecell{Employ data spectrum \\ to design backdoor}\\
            \midrule
            NFTA (2023) \cite{DBLP:journals/ijis/ChenYZW23} & $\bullet$ & - & - & $\bullet$ & - & - & General GNN & \makecell{Improve performance by \\ smoothing out triggers}\\
            \midrule
            Link-Backdoor (2023) \cite{10087329} & - & - & $\bullet$ & $\bullet$ & - & $\bullet$ & General GNN & \makecell{Extend backdoor to\\ edge-level task}\\
            \midrule
            Link Prediction Attack (2024) \cite{link-bkd-2024} & - & - & $\bullet$ & $\bullet$ & - & - & General GNN & \makecell{Improve edge backdoor \\ triggers}\\
            \midrule
            MLGB (2024) \cite{WANG2024110449} & - & $\bullet$ & - & $\bullet$ & $\bullet$ & $\bullet$ & General GNN & 
            \makecell{Introduce \textit{N-to-N}
            multi-target \\ backdoor attack}\\
            \midrule
            Multi-Target Backdoor (2023) \cite{10.1145/3576915.3624387} & - & $\bullet$ & - & $\bullet$ & - & - & General GNN & 
            \makecell{Introduce \textit{1-to-N}\&\textit{N-to-1}
             \\ multi-target backdoor}\\
            \bottomrule
        \end{tabular}
    }
\end{table*}

\subsection{Framework of GNN backdoor}
For GNN model, adversaries typically resort to data-poisoning on the training data to implant backdoors.
This manipulation will cause the well-trained GNN to produce adversary-designated outputs when input trigger-embedded graph data during deployment.
Nevertheless, for clean inputs (unpoisonous or unattacked), the model will maintain its accuracy.

To be specific, implanting and activating the backdoor through poisoning generally requires three steps: (\textit{i}) poisoned data generation; (\textit{ii}) backdoor model training; (\textit{iii}) backdoor activation.

In the first step, adversaries select a subset $\mathcal{D}_t$ from the training data $\mathcal{D}$ based on poisoning rate $\gamma$ (proportion of the selected), wherein specific triggers $\Delta$ (\textit{e.g.}, subgraphs, nodes, edges, or global properties) are subsequently inserted into $\mathcal{D}_t$ and labels of samples in $\mathcal{D}_t$ are altered to match the adversary-specified target class. This data-poisoning process can be expressed as

\begin{equation}
\mathcal{D}_t \sim \mathcal{D}, \quad |\mathcal{D}_t| = \gamma|\mathcal{D}|,
\end{equation}

\begin{equation}
\mathcal{D}_t = \{ (x_i=x_i+\Delta, y_t) \, | \, (x_i, y_i) \in \mathcal{D}_t \}.
\end{equation}

Following data-poisoning, the poisoned dataset $\mathcal{D}$ is utilized for GNN training (\textit{i.e.}, the second step). This process can either follow the standard procedure or be subject to adversary influence (\textit{e.g.}, modifying the training pipeline or loss). This is represented as
\begin{equation}
\text{GNN}_{\theta} = \underset{(x_i, y_i) \in \mathcal{D}}{\text{argmin}_{\theta}} \mathcal{L}(\text{GNN}(x_i; \theta), y_i),
\end{equation}
where loss $\mathcal{L}$ and the $\theta$ optimization could be controlled by adversaries.

After training is completed, GNN gets infected (backdoored). It generates adversary-specified results $y_t$ when given trigger-embedded samples during testing, while retaining prediction accuracy for normal data (\textit{i.e.}, backdoor activation):
\begin{equation}
\text{GNN}(x_i + \Delta) = y_t, \quad \text{GNN}(x_i) = y_i.
\end{equation}

\subsection{Attack Methodologies}
Based on the classification of backdoor attacks studies on GNN outlined at the beginning of this section, we will elaborate on them in two parts: (\textit{i}) adaptability-expanding research; (\textit{ii}) effectiveness-improving research. The comprehensive summary of backdoor attacks is shown in Tab. \ref{bkd-sum}.

\subsubsection{Adaptability-Expanding Research}
These studies center on the deployment of GNN backdoors across diverse scenarios, including new graph learning paradigms and practical real-world applications.

\textbf{\textit{Pioneering Works.}}
Zhang et al. initially investigated the feasibility of embedding backdoors within GNNs \cite{10.1145/3450569.3463560}.
This was achieved through using subgraphs as triggers, which were strategically inserted into the graph structures of a randomly selected subset of training samples, and simultaneously, altering the source labels of these samples to the target class (\textit{e.g.}, data-poisoning).
The poisoned training set (including poisoned subset and clean subset) was then utilized for GNN learning, and this led to the trained model becoming infected (backdoored), resultantly causing misclassification of trigger-embedded graph inputs into the adversary's predetermined target class. Notably, the model still maintained its accuracy in classifying regular graph samples amidst this backdoor affection.

This study further explores the impact of different categories of subgraph triggers on the effectiveness of the backdoor. It introduces two types of trigger designs: fixed-pattern subgraphs and generated-pattern subgraphs. While the fixed-pattern subgraph mode (\textit{e.g.}, complete subgraph) is relatively singular and easier to detect and filter, a generation algorithm (\textit{e.g.}, preferential attachment model) is employed to create diverse subgraphs, enhancing the stealthiness of the attack. Moreover, \cite{DBLP:conf/colcom/ShengCCK21} presented a similar attack scheme and optimizes the selection of target subgraph for trigger insertion by degree centrality and closeness centrality.

Another pioneering research in GNN backdoor is Graph Trojaning Attack (GTA) \cite{272256}. 
It also leverages data-poisoning to infect the training dataset to embed backdoor. 
However, in contrast to Subgraph Backdoor, it differs in two key aspects: trigger design and model training.
Firstly, it doesn't employ fixed models of trigger generation for backdoor implantation. Instead, it utilizes a dynamically-trained backdoor trigger generator to craft subgraph trigger tailored to individual input graph structure. 
Secondly, it introduces a bi-level optimization training paradigm. This involves concurrently training the model and the trigger generator, which not only confirms the normal training of the model but also enforces the generator to create corresponding trigger subgraphs based on the features of input graph structures, rather than relying on fixed subgraph-generation model or algorithm.

Besides backdooring graph classification, GTA can be extended to node-level tasks. For these attacks, GTA aims to classify all nodes within \textit{K} hops from the trigger subgraph into the specified target class, where \textit{K} represents the hidden layer numbers of GNN. Recalling GNN properties, a node exerts its influence on others at most \textit{K} hops away, and hence the trigger subgraph can impact nodes within its \textit K hops. Accordingly, when training GNN, loss functions are designed to minimize the loss between these nodes and the target class, while simultaneously facilitating regular training for other nodes, which results in nodes within \textit{K} hops from the trigger in the attacked graph being classified as target during testing.

\textbf{\textit{Federated Graph Learning Backdoor.}}
Federated Graph Learning is a decentralized machine learning paradigm where multiple participants collaborate to train a model without sharing raw data.
In each training epoch, individual clients retrieve the global model from the server and perform model training based on local data. Following this, the updated local GNN parameters are uploaded back to the server and aggregated into the global model.
This approach preserves privacy by keeping the training data locally. However, it also introduces the risk of data-poisoning backdoors.

Xu et al. first explored the possibility of implanting backdoors into this paradigm and presented two types of data-poisoning-based strategies: Distributed Backdoor Attack (DBA) and Centralized Backdoor Attack (CBA) \cite{10.1145/3564625.3567999, 10.1145/3633206}. 
\begin{enumerate}
    \item Distributed Backdoor Attack. 
    A global graph trigger is generated via graph generation algorithm, which is then decomposed into diverse local subgraph triggers allocated to multiple malicious clients for training their respective locally backdoored models. Then, the updated model parameters of these malicious and normal clients are transmitted to the central server and aggregated, which results in backdoor implanting into the global model.

    \item Centralized Backdoor Attack.
    Different from DBA, CBA employs the global graph trigger to poison only one malicious client and backdoors the entire global model through model parameter updates from the single malicious client.
    
\end{enumerate}


Liu et al. extended the backdoor to node classification scenarios via a stated local loss function \cite{DBLP:journals/corr/abs-2306-10351}. Additionally, they discussed various factors influencing the backdoor performance in federated graph learning (\textit{e.g.}, data distribution, adversary client number, trigger size, location, poisoning rate, etc.).

\textbf{\textit{Contrastive Graph Learning Backdoor.}}
Graph Contrastive Learning (GCL) is designed to learn graph representations in the absence of model supervision. It is achieved by randomly augmenting the input graph into two views. To be specific, a GNN encoder will map graph nodes in various views into node embeddings, and the encoder is trained to minimize the classical InfoNCE objective.

To implant backdoor into the GCL framework, Zhang et al. first looked into the viability and proposed three methods according to the attack scenarios: (\textit{i}) data-poisoning attack; (\textit{ii}) set-crafting attack; (\textit{iii}) natural backdoor attack \cite{pmlr-v202-zhang23e}.

\begin{enumerate}
    \item Data-poisoning Attack. 
    This attack confines adversaries to control the training data collection process of GCL.

    One core concept of GCL involves training to maximize the similarity between embeddings of the same graph node across different mapped views. Hence, by selecting a subset of nodes from the target class in the graph and connecting them to nodes with triggers, certain connections may disappear in some views after mapping through the encoder (nodes won't aggregate trigger node information), while others will remain (nodes will aggregate trigger node information). Consequently, when calculating the loss, the encoder training aims will maximize the similarity between embeddings with trigger information and those without, hence achieving the effect of a backdoor.

    \item Set-crafting Attack.
    This attack aims to implant a hidden backdoor into a previously clean pre-trained encoder, enabling any subsequent classifier built upon it to inherit the covert backdoor logic.
    It is achieved by fine-tuning the encoder, which involves employing two designed loss functions. For normal node samples, the objective is to maximize the similarity between the node embeddings of the backdoored encoder and the clean encoder. Simultaneously, for trigger-embedded nodes, the aim of loss is to maximize the similarity between their embeddings and those mapped from the target class node data.

    \item Natural Backdoor Attack.
    The attack condition of the natural backdoor adversary closely resembles that of the crafting adversary, with the key distinction being that the natural backdoor adversary is constrained from altering the clean encoder. 
    
\end{enumerate}

\textbf{\textit{Physical Graph System Backdoor.}}
This type of attack aims to implant backdoors in practical GNN physical application systems. It achieves the backdoor objective by designing corresponding subgraph triggers based on actual hardware conditions and modifying the model's predicted outcomes, resulting in the backdoored systems in response to trigger-embedded inputs.

Research \cite{10114622} initially investigated the GNN backdoor in integrated circuits, utilizing subcircuits as triggers to infect the circuit dataset used for training. The embedded subcircuits do not affect the performance of the original circuit and can pass functional convergence tests. The backdoored GNN physical system classifies the circuit with triggers as harmful. Experimental results based on Hardware Trojan Detection Systems and Intellectual Property Piracy Detection Systems demonstrate that the proposed backdoor scheme exhibits excellent performance.

\subsubsection{Effectiveness-Improving Research.}
These attacks generally enhance the efficiency, concealment, and robustness of backdoors, increasing the difficulty of defense detection and filtering during the implementation.

\textbf{\textit{Explainable Graph Backdoor.}}
Designing the triggers is a key aspect of implanting backdoors.
However, selecting the optimal trigger implantation positions and determining the associated graph or node attribute values of the trigger to enhance backdoor efficacy remain unexplored.

Xu et al. proposed an Explainability-based Backdoor Attack, exploiting node features with either the strongest or weakest characteristics to design subgraph triggers for poisoning the training data and embedding backdoors \cite{10.1145/3468218.3469046}.
Executing attacks using data with prominent features makes the GNN model efficiently learn the corresponding trigger features, and facilitates more effective backdoor implantation. Conversely, launching backdoor attacks using data with weaker features minimally affects the feature learning process for normal samples, rendering the attack more covert and less prone to detection.

Specifically, for graph classification tasks, it initially trains a regular GNN model with clean data, and subsequently utilizes the explainability tool, \textit{GNNExplainner}, to identify \textit{t} least important nodes (where \textit{t} represents the number of subgraph trigger nodes). These nodes are then replaced with the predefined subgraph trigger, and the corresponding data labels of graph are modified to the target class, namely poisoning the training data and leading to the backdoor implantation within the trained model.

It also extends the attack to node classification tasks. 
Similarly, it first well-trains a node classification GNN and leverages \textit{GNNExplainner} to identify the least representative \textit{n} dimensional features in node vector.
Next, it substitutes these identified \textit{n} features of the victim nodes with trigger feature values, and alters the source labels to the target class, resulting in a backdoored GNN after training.

Zheng et al. researched the distribution patterns in statistical terms between trigger subgraphs and normal subgraphs \cite{10108961}. Their findings indicated that utilizing subgraphs divergent from the distribution of normal samples as triggers resulted in a notable enhancement in attack performance. Moreover, based on this phenomenon, they proposed a motif-based backdoor poisoning technique. 
This method employs available data to select significantly skewed subgraph distributions as motif triggers and applies node importance and subscore to determine the optimal trigger injecting position with the highest values. Then, the adversary can poison the training data with the designed trigger and backdoor the associated GNN models via learning.

Chen et al. introduced a general GNN backdoor tailored for node-level tasks \cite{10063438}. 
Employing a trigger generator, the attack generates constrained node features specifically designed for poisoned targets, with both the GNN model and the generator being jointly trained via the poisonous dataset, which resultantly creates a backdoored GNN.
Furthermore, to reduce the impact of other nodes, it selectively eliminates critical edges of targets, identified through explainability analysis. Additionally, the trigger feature dimensions crucial for targets are prioritized.

Research \cite{WANG2024106097} proposed an explanatory subgraph-based poisoning attack. The subgraph incorporates graph structure and node feature information relevant to the target label, and this strengthens the association between the subgraph and the target while minimally altering the original data's structure, and preserving its statistical properties.
Also, the adversary will modify the explanation information of the graph, which makes the change of the graph more covert and reasonable.

\textbf{\textit{Unnoticeable Graph Backdoor.}}
Current GNN backdoors require a significant attack budget for an effective attack assault, and hence, the injected triggers could be easily detected and pruned. 

Dai et al. developed imperceptible trojan backdoor attacks under limited attack budgets \cite{10.1145/3543507.3583392}. 
Utilizing a trigger generator, the generation of infective samples ensures a sufficiently high similarity with poisoned samples from the target class data. Additionally, to make the model maintain the training accuracy on normal samples, a bi-optimization loss is designed to train both trigger generator and GNN. Furthermore, for the selection of the target for the attack, the method leverages the obtained node embeddings for clustering and chooses the most representative nodes as targets to assure backdoor efficiency.

To achieve attack concealment in graphs, study \cite{dai2024cleangraph} developed a clean-graph backdoor attack. 
This attack refrains from modifying the graph's features or structure and does not introduce malicious nodes and edges prior to training. Instead, adversaries solely alter the labels of victim nodes possessing adversary-desired distinctive features. 
By doing so, they ensure that the GNN training process leads the model to produce target outcomes when presented with input data exhibiting these features.

\textbf{\textit{Clean-Label Graph Backdoor.}}
Clean-label attack refers to a form of poisoning attack in which no modifications are made to the labels of poisoned data. Current GNN backdoor attacks achieved through data-poisoning typically contain altering the source labels of infected data, but these modified labels often diverge significantly from the actual semantic content of the original data, making them more susceptible to detection and filtration by defense mechanisms. Accordingly, clean-label attacks exhibit augmented stealthiness as they avoid such overt label modifications.

Xu et al. initially delved into the feasibility of implementing clean-label backdoor attacks in GNNs \cite{10.1145/3548606.3563531}. It samples victim targets within the training dataset pertaining to the target class, inserting triggers within them, and then training the GNN model using both poisoned and regular data.
Since the infected data belongs to the target class, it suffices to embed triggers in these samples to enable the GNN to learn trigger features without label revision.
During the testing phase, the GNN will classify trigger-contained inputs as belonging to the target class.
To enhance this method, Xing et al. select robust features of nodes with larger degrees as triggers \cite{xing2023cleanlabel}.
These features are typically strongly associated with their respective labels. Utilizing them as triggers facilitates the model in establishing a strong correlation between the trigger and the target label.

Yang et al. evaluated the potential of clean-label attacks for node classification \cite{percba}. To realize the backdoor effect through poisoning without label modification, they embedded triggers within the target node data and introduced perturbations therein. This action causes the decision boundary of the target category to lean toward infected data during training, and leads the model to predict attacked data as the target class during the testing phase. Simultaneously, to enhance the normal performance robustness of GNN, they also proposed a contrastive poisoning strategy to slightly poison normal samples.

\textbf{\textit{Transferable Graph Backdoor.}}
The transferable backdoor aims to enhance the transferability of backdoors based on models, maximizing the success rate of attacks on different types of GNNs. This attack allows the adversary to design the corresponding data-poisoning strategy without provided knowledge regarding the victim model. 

The transferable GNN backdoor attack is proposed in \cite{10.1145/3545948.3545976}, utilizing available real-world data to create triggers for poisoning and implanting a backdoor in a shadow GNN model via training. The mode of subgraph triggers is optimized during training using loss function derivatives to maximize the attack success rate, and it could be applied in poisoning various downstream task GNN models (\textit{i.e.}, transferability). Additionally, due to the harsh computing of optimizing graph structure derivatives, a solution strategy for graph structure perturbations was also presented in the study.

\textbf{\textit{Spectral Graph Backdoor.}}
Most GNN backdoors define graph trigger exclusively within the spatial domain, thus limiting the effectiveness of attacks. 

Inspired by frequency domain graph theory and frequency-based backdoor attack methods, Zhao et al. proposed an effective spectral graph backdoor attack method \cite{10318195}. 
It injects subtle trigger signals into the crucial frequency bands of important node features to poison training samples and implant backdoors within victim GNN. For the trigger, specifically, it first extracts the ego graph constructed from the attributes of the neighboring nodes of the target node and performs spectrum decomposition on this ego graph. Then, the adversaries inject a small trigger signal into the spectrum graph after the spectrum decomposition. The injection is done at the important frequency bands of the important features, to guarantee an efficient backdoor effect.

\textbf{\textit{Graph Node Feature Backdoor.}}
A novel backdoor approach is proposed based on graph node features, aiming to infect nodes and optimize the smoothness of graph features to generate an infected graph for backdooring GNNs \cite{DBLP:journals/ijis/ChenYZW23}. 
Concretely, the attack employs a mask to conceal node feature data and derive features. Following that, a graph feature optimization function is utilized to modify the graph structure, resulting in the smoothest infected graph containing the trigger. 
The features of the graph are smoother, minimizing the impact on feature learning from the original data. The backdoor keeps better preservation of the model's accuracy on normal samples while ensuring the robustness of the backdoor performance.

\textbf{\textit{Graph Link Backdoor.}}
Link prediction is an essential task in GNN, primarily targeting at predicting links (edges or connections) that are not yet present but may appear in the future within a graph.
Zheng et al. initially analyzed the feasibility of implementing backdoors in such tasks, exploiting poisoning to backdoor GNN \cite{10087329}.
It first designates a specified subgraph structure as the trigger, then optimizes it using gradient information from the prediction model, and finally poisons the training data with the optimized trigger to backdoor the victim GNN.
Dai et al. further improved the trigger by leveraging nodes as the trigger. During training, edges between nodes connected to this trigger are misclassified into an adversary-specified target \cite{link-bkd-2024}.

\textbf{\textit{Multi-Target Graph Backdoor.}}
The multi-target backdoor attack refers to embedding the backdoor that can output multiple possible results based on the adversary's choice, rather than single one.

Study \cite{WANG2024110449} reviewed the possibility of implementing such attacks within graph learning framework. 
First, explainability techniques are applied to determine multi-trigger locations and identify poisoned graph nodes according to these locations. 
Next, a one-hot encoding mechanism is used to encode the poisoned nodes and multiple target labels. 
Finally, one carefully designed loss function, coupled with a two-tier-optimization mechanism, is leveraged to train the multi-trigger generation model and victim GNN, resulting in the GNN backdoored that responds to multi-triggers after learning.

Xu et al. implemented multi-target backdoors through multi-target data-poisoning \cite{10.1145/3576915.3624387}. 
The attack primarily centers on two backdoor strategies: \textit{One-to-N}, in which distinct numerical values of several feature dimensions regulate various target outcomes, and \textit{N-to-One}, whereby when all the triggers are satisfied, the backdoor could be activated.

\begin{table*}[t]
    \label{bkd-def-sum}
    \caption{Summary of Graph Neural Backdoor Defense Strategies}
    \resizebox{\textwidth}{!}{
        \begin{tabular}{ccccccccc}
            \toprule
            \multirow{2}*{\textbf{Defense Strategies}}& \multicolumn{3}{c}{\textbf{Defense Level}} & \multicolumn{3}{c}{\textbf{Defender Capability}} & \multirow{2}*{\textbf{Defense Function}} & \multirow{2}*{\textbf{Research Novelty}}\\
            \cmidrule(lr){2-4} \cmidrule(lr){5-7}
              & Graph & Node & Edge & Sample & Model & Training & &\\
            \midrule
            BloGBaD (2024) \cite{10.1007/978-981-97-0808-6_10} & $\bullet$ & - & - & $\bullet$ & - & $\bullet$ &      \parbox[m]{2.8cm}{
                \raggedright
                $\diamond$ Backdoor detection \\
                $\diamond$ Sample filtration \\
                $\diamond$ Backdoor mitigation
            } & \makecell{Employ sample distributions \\ to identify infected graphs}\\
            \midrule
            Explainability Defense (2022) \cite{bkd-df-2} & $\bullet$ & - & - & $\bullet$ & $\bullet$ & - & \parbox[m]{2.8cm}{
            \raggedright
            $\diamond$ Backdoor detection \\
            $\diamond$ Sample filtration
        } & \makecell{Leverage explainability tools to \\ detect poisonous graphs}
         \\
            \midrule
            Securing GNN (2024) \cite{downer2024securing}  & $\bullet$ & -& - & $\bullet$ & $\bullet$ & - & \parbox[t]{2.8cm}{
            $\diamond$ Backdoor detection
        } & \makecell{Use graph features to \\ detect sample anomalies}
             \\
            \bottomrule
        \end{tabular}
    }
\end{table*}

\subsection{Defense Methodologies}
In response to the threat of GNN backdoor attacks, researchers have begun focusing on developing effective defense strategies. The comprehensive summary of GNN backdoor defense studies is displayed in Tab. \ref{bkd-def-sum}.

Current defense mechanisms against backdoors in GNN primarily concentrate on sample filtering or detection. This entails analyzing the features of test graphs to identify anomalies or to determine if the model is infected by the backdoor, and correspondingly
 filtering out the anomalous parts to prevent its activations.

Research \cite{10.1007/978-981-97-0808-6_10} proposed a defense method based on data filtering. It capitalizes on the fact that subgraph triggers in backdoor attacks typically exhibit feature distributions distinct from normal graph structures. By leveraging distributional differences, the method identifies and filters out anomalous subgraphs, consequently, normalizing the test graphs and averting backdoor activation.
Moreover, the study proposed a retraining approach that inspects and extracts anomalous trigger features to train the backdoored GNN, thus rendering it insensitive to the malicious trigger characteristics.

Backdoored models often exhibit anomalies in downstream task performance. Therefore, Jiang et al. employed explainability tools to detect and filter anomalies originating from test graphs \cite{bkd-df-2}. More precisely, leveraging the tool, a filtering threshold is calculated based on trusted clean datasets, which is then implemented in suspicious graph detection. 
Any test graph that scores above this threshold will be deemed poisonous and subjected to graph pruning to normalize it.

Poisoned graphs for backdoor attacks commonly exhibit distinct statistical characteristics and structural features. Derived from this observation, Downer et al. assessed various aspects including Prediction Confidence, Connectivity, Subgraph Node Degree Variance, Elbow, and Curvature, and these factors were then analyzed to identify harmful segments within the test graph to detect backdoor triggers \cite{downer2024securing}.

\section{Applications and Challenges}
Although the studies of GNN backdoors typically center on their threats for security vulnerabilities, it's crucial to recognize the untapped potential these mechanisms offer in beneficial domains. 
Beyond their capacity for malicious intent, GNN backdoors also present opportunities for enhancing the resilience, privacy, and trustworthiness of AI systems. 
This section endeavors to elucidate the practical applications of GNN backdoors, revealing the possible (positive) applications across various domains.


\subsection{Model Intellectual Property Protection}
As AI models find widespread application across various industries, the associated development costs and complexity have risen substantially, along with their commercial value. Concurrently, the issues of model misuse and theft have become more prevalent, prompting greater attention from industry and legal sectors toward intellectual property (IP) protection.

The activation of GNN backdoors typically requires the input to contain a trigger, while remaining dormant in other cases.
If a certain type of backdoor could be embedded during the design and training of the model, such that GNN outputs correct predictions only when the input contains the trigger while providing specific erroneous results otherwise, then the trigger can be treated as a special token to control the model's input-output behavior, thereby achieving the goal of model IP protection.
Based on this core idea, backdoor mechanisms could be applied in IP protection.
The IP protection studies via backdoor can primarily be divided into three categories: (\textit{i}) watermark verification; (\textit{ii}) controlling authorization; (\textit{iii}) model stealing countermeasures.

\textbf{\textit{Watermarks Verification.}}
This mechanism mainly utilizes watermarks as backdoor triggers to implant into the model. 
The main idea is to employ watermark (\textit{i.e.}, known only to legal users and administrators) to poison clean data without label altering, and optimize the distance between the predictions of poisonous data and the ground truth during training while randomizing the output results of others.
Since the watermark is set during the training process, only administrators and users have access to it, and the well-trained model can only predicts with high-accuracy performance when the input data contains the watermark \cite{sun2023deep, 10097580, 10.1145/3196494.3196550}.

To protect graph learning IP in downstream tasks such as recommendation systems, molecular classification systems, and intelligent traffic guidance systems, MLaaS providers (\textit{e.g.}, Google and Amazon) need to set user-specific graph watermarks (\textit{e.g.}, subgraphs, nodes, or global features) according to practical requirements before the training phase, and then utilize these to poison the data without label modification and train GNN, finally embedding the watermark \cite{10190545}.

To further optimize the backdoor-driven GNN watermark to better adapt to graph learning, it is necessary to concurrently develop:
\begin{enumerate}
    \item Dataset Watermark. 
    Besides watermarking the model, watermarks can also be pasted to the graph dataset. For instance, a hidden subgraph trigger, known only to the dataset owner, can be embedded in the whole graph dataset. When this dataset is leveraged for GNN training, it leaves a backdoor in the trained model, and the trigger can verify the backdoor, which enables the owner to determine dataset misusing\cite{li2020opensourced}.

    \item Watermark Verification Mechanism. 
    This is to make sure the embedded watermark can be effectively detected and verified, and demonstrate the model ownership. 
    Verification can be achieved through the style of triggers, the behavior of the model under specific inputs, or differential privacy measures \cite{wang2022nontransferable, pei2022practical, info11020110}.
    
    \item Watermark Removal Countermeasures. 
    Watermark removal refers to detecting and eradicating the embedded watermark within GNN by adversaries.
    The defense schemes include introducing adversarial loss during training to render GNN watermark more robust \cite{tramèr2017space, NEURIPS2019_e2c420d9}.
    
    
\end{enumerate}

\textbf{\textit{Controlling Authorization.}}
This mechanism is inspired by access control, where GNN model produces normal outputs only for data with valid embedded user tokens, while outputting anomalies for all others. 

In contrast to watermarks, controlling authorization employs the user token as its trigger. Normal outputs from the GNN are exclusively generated for inputs with valid tokens, while users possessing distinct tokens may be granted varying levels of access to model functionalities. 
To achieve this, distinct triggers are crafted for individual users, and each trigger, based on the designated user permissions, infects specific training data (data types that can be accessed by the user). Consequently, the trained GNN will generate outputs solely for user input embedded with valid tokens, with the output scope restricted to the areas the user is authorized to access \cite{xue2023turn}.

To enhance backdoor-based GNN controlling authorization, the subsequent domains necessitate further development:
\begin{enumerate}
    \item Access Control Policy. 
    Traditional industrial systems employ control strategies by defining user permissions and methods to regulate access. To implement role-based access control (RBAC), attribute-based access control (ABAC), and rule-based access control (RuBAC) in GNN-based systems using solely backdoor techniques, MLaaS company or system administrators should design appropriate poisoning strategies to warrant that the GNN makes outputs consistent with the user's role and access permissions based on various input samples associated with different user tokens. Furthermore, scalable authorization strategies should be considered to warrant that the control system can accommodate a large user base and dynamically add or remove users \cite{312842, 8968396, 5462174, 10.1145/501978.501980}.


    \item Token Privacy Protection. 
    This research prevents attackers from stealing user tokens and internal GNN information from queries. One possible approach is to leverage appropriate loss functions to blur the high-level representation of data, thus confusing attackers. Another feasible strategy is data anonymization or introducing noise (\textit{e.g.}, differential privacy) \cite{Doan_2021_ICCV, 9724389, 9870671}. 
      
\end{enumerate}

\textbf{\textit{Model Stealing Countermeasures.}}
Model stealing (or extraction) attack refers to an AI system attack where an attacker attempts to reverse-engineer or steal the target GNN model's parameters, architecture, or training set by querying the target model with carefully crafted inputs and observing its responses. Current attack allows adversaries to replicate a surrogate GNN with comparable accuracy to the original one, which steals the copyright of MLaaS providers and avoids payment of the usage fees.

Traditional watermarks are ineffective in safeguarding model copyrights in this attack scenario. 
This limitation arises from the separation of model parameters related to the primary task (classification) and the watermark task (backdoor), which occurs due to overfitting. As a result, when adversaries attempt to extract the primary task's functionality, the watermark, as a separately-distinct task within the model, may not be transferred to the surrogate model.

To bridge this gap, the corresponding defense strategy couples the parameters of the primary task with the backdoor task during training via the restricted loss function. Study \cite{272262} initially identified this issue and addressed it by employing Soft Nearest Neighbor Loss during the backdoor training process. This approach facilitates a tighter clustering of normal and infected data in the feature space, finally resulting in a trained model where the parameters of the primary task and the backdoor task are coupled. Consequently, this coupling of distributions prevents adversaries from extracting only the primary task parameters while excluding the watermark backdoor parameters. 

To efficiently deploy model stealing countermeasures in graph learning, several fields can be considered:
\begin{enumerate}

    \item Real-time Watermark.
    Implementing dynamic watermark embedding mechanisms can make the watermark information evolve (since real-world graph data often dynamically changes over time), and increase the difficulty for adversaries to detect or remove the watermark. It designs algorithms that automatically update and maintain the watermark, and warrants the effectiveness of the watermark as the graph topology changes. 
    
    \cite{real-bkd-detect}.

    \item Federated Verification.
    Graph data is typically owned by multiple distributed entities (\textit{e.g.}, social network users, organizations, or devices), and they may be unwilling to centralize data due to privacy concerns. The federated learning (FL) allows model training and proceeds in such scenarios, and hence, combining entangled watermarking techniques with FL will make the watermark distributed across multiple models, and thus enhance its resistance to extraction attacks. 
    \cite{9603498, 9658998, 10.1145/3630636}.

\end{enumerate}

\subsection{Defending Against Adversarial Attacks}
Adversarial attacks exploit vulnerabilities in AI models, resulting in inaccurate outputs or misdirection, rather than intended right classification results. This attack method commonly utilizes adversarial sample generation algorithms to introduce nuanced perturbations for inputs to induce erroneous predictions by the model.

In adversarial attacks, the perturbation added to the original input data that causes the model to perform incorrectly is actually a type of noise. If we can leverage the perturbation as the trigger to poison the training dataset and modify the ground truth of the infected data as a specific category set by the defender (unrelated classification outputs, \textit{e.g.}, errors), then after the model training and deployment, any adversarial attacks attempted by the adversary will lead to one same defender-specified erroneous result from the model.

\textbf{\textit{Honeypot Defense Strategy.}} 
The above concept was implemented in study \cite{10.1145/3474370.3485655}, wherein perturbations were utilized as triggers to infect the training data, followed by fine-tuning the model to incorporate defensive-purpose backdoors. 
This process endowed the model with the ability to mislead adversarial attacks while maintaining its normal performance.
Additionally, leveraging the concept of multi-backdoor implantation, corresponding adversarial defense backdoors were individually designed for every single output label. 
Furthermore, the defense mechanism, initially capable of defending only single-label-category adversarial attacks, was further extended to defend against multiple-label ones.
Researchers name this defense mechanism as ``\textit{honeypot}," with the employed backdoor referred to as a ``\textit{trapdoor}," owing to its ability to attract all adversarial attack samples and channel them into a designated useless category.

To further optimize this honeypot defense method in graph learning, the following aspects can be considered for improvement:

\begin{enumerate}
    \item Honeypot Detection Countermeasures. Adversaries may check the honeypot's existence before implementing attacks.
    To prevent such detection, the defender can adopt the following technical strategies: (\textit{i}) employ diverse trapdoor designs with optimized triggers and randomization strategies to reduce predictability; (\textit{ii}) conceal trapdoor behavior to minimize deviation from normal model behavior and avoid obvious detection features (\textit{e.g.}, output random results instead of specific target); (\textit{iii}) regularly update the trapdoor settings dynamically to prevent static analysis from identifying patterns; (\textit{iv}) use obfuscation to further enhance the concealment of trapdoors (\textit{e.g.}, noise injection or data sanitization) \cite{Tao_2022_CVPR, 9186317, Yu_2023_CVPR, 5958021}.


    \item Honeypot Transferability. The research mainly focuses on transferring and applying honeytrap from one domain to another while maintaining its robustness and effectiveness
    \cite{Lee_Kim_Lee_Yoon_2017, 9684927, 9878092}.
    
\end{enumerate}

\subsection{Machine Unlearning Verification}
As a company providing MLaaS, it typically collects datasets from various users to train models on its platform. 
However, in certain situations, users may request to withdraw their uploaded or shared data and demand the company to erase any learned content from that dataset within the AI models (\textit{e.g.}, due to copyright expiration or termination of collaboration). For this purpose, MLaaS companies might employ unlearning methods. Nevertheless, verifying whether the relevant content has been successfully removed poses a significant challenge, as the companies could maliciously claim to use data with similar distributions while continuing to train on the original set.

To address this issue, the backdoor mechanism is introduced into the verification. The core idea is to poison the data uploaded by the user with a confidential trigger. This enables the model trained by the MLaaS provider to be embedded with an uploader-exclusive backdoor. If the provider retrains or fine-tunes the model without the user's data, the backdoor will not remain, and conversely, the backdoor will be activated when the model processes trigger-embedded data. Since the trigger pattern is owned only by the user, it can be utilized to verify whether the user-data-relevant information has truly been revoked \cite{sommer2020probabilistic}.

Although backdoor-based unlearning verification has been proven to be efficient in handling linear data, unique challenges still exist when dealing with graph learning. Traditional machine learning methods often encounter limitations in processing graph data due to the topological structure of the graph and the representation of node attributes. 
The following are some research directions that can be explored:

\begin{enumerate}
    \item Invisible Graph Trigger. 
    If triggers within the uploaded data are noticeably conspicuous, MLaaS providers may undertake countermeasures to prevent backdoor implantation. 
    To prevent this, the research can focus on designing minimal graph perturbations to serve as triggers, and poisoning training data to launch backdoor attacks, so as to effectively evade detection.

    \cite{li2020invisible, 9747582, 9747888582}.

    \item Improve Verification Security. 
    This is aimed at safeguarding against the leakage of backdoor information during the verification, integrating backdoor verification techniques with differential privacy protection, homomorphic encryption, trusted execution environments, secure multi-party computation, etc.
    \cite{10.1145/3437880.3460401, li2021rethinking, 10.1007/978-3-030-58607-2_11}.

\end{enumerate}

\section{Future Directions}
Although several research efforts have been devoted to investigating the nature of graph backdoors, the field still faces numerous unresolved issues and should be further dug. 
This section will delve into the GNN-backdoor-related future directions, aiming to provide novel insights and approaches for addressing backdoor security concerns in graph learning and advancing the field's development and innovation.

Future research directions could be primarily divided into three main parts: 

\begin{itemize}
  \item \textbf{Backdoor Optimization.} 
  This part will focus on further improving and optimizing existing attack methods to enhance their attack efficiency to help understand backdoor characteristics and bust defense side research. 
  
  \item \textbf{Applicability Extension.}
  It extends GNN backdoors to other complex learning scenarios, which include but are not limited to domain adaptation, meta-learning, multi-modal learning, etc., further expanding the backdoor applicability in practical applications.

  \item \textbf{Countermeasure Development.}
  Beyond enhancing the accuracy of the defense methods, it is crucial to formulate defense strategies tailored to various threat models (\textit{i.e.}, different adversary and defender abilities).
  
\end{itemize}

A brief summary for the future directions of GNN backdoor is shown in Tab. \ref{future-direction}. 

\begin{table*}[h]
\caption{Summary of Future Directions for Graph Neural Backdoors}
\label{future-direction}
\centering
\renewcommand{\arraystretch}{1.5} 
\begin{tabular*}{\hsize}{@{\extracolsep{\fill}} c m{4.5cm} m{8.3cm}}
\toprule
\textbf{Direction} & \textbf{Research Focus} & \textbf{Specific Topic} \\
\midrule
\multirow{3}{*}{\centering Backdoor Optimization} 
  & \parbox[c][3em][c]{4.5cm}{\centering Graph Semantic Backdoor} 
  & \parbox[c][3em][c]{8.3cm}{
            $\diamond$ Adaptive trigger-optimization\\
            $\diamond$ Hide semantic features of trigger
        } \\
\cline{2-3}
  & \parbox[c][3em][c]{4.5cm}{\centering Black-Box Graph Backdoor} 
  & \parbox[c][3em][c]{8.3cm}{
            $\diamond$ Inverse engineering-based approach \\
            $\diamond$ Surrogate model-based data-poisoning
        } \\
\cline{2-3}
  & \parbox[c][2em][c]{4.5cm}{\centering Dormant Graph Backdoor} 
  & \parbox[c][2em][c]{8.3cm}{$\diamond$ To make GNN backdoor triggerable only after downstream fine-tuning} \\
  \cline{2-3}
  & \parbox[c][2em][c]{4.5cm}{\centering Untargeted Graph Backdoor} 
  & \parbox[c][2em][c]{8.3cm}{$\diamond$ To implant target-unspecified graph neural backdoors} \\
\cmidrule(lr){1-3}
\multirow{5}{*}{\centering Applicability Extension} 
  & \parbox[c][1em][c]{4.5cm}{\centering Generative Graph Backdoor} 
  & \parbox[c][1em][c]{8.3cm}{$\diamond$ To implant backdoor into generative graph neural models} \\
\cline{2-3}
  & \parbox[c][2em][c]{4.5cm}{\centering Few-Shot Graph Backdoor} 
  & \parbox[c][2em][c]{8.3cm}{$\diamond$ To implant backdoor into few-shot-based GNN models} \\
\cline{2-3}
  & \parbox[c][6.5em][c]{4.5cm}{\centering Graph Large Model Backdoor} 
  & \parbox[c][6.5em][c]{8.cm}{
            $\diamond$ Fine-tuning backdoor embedding \\
            $\diamond$ Distributed backdoor embedding \\
            $\diamond$ Data augmentation backdoor embedding \\
            $\diamond$ Multi-modal Graph backdoor embedding
        } \\
\cline{2-3}
  & \parbox[c][2em][c]{4.5cm}{\centering Parameter Modification Backdoor} 
  & \parbox[c][2em][c]{8.3cm}{$\diamond$ To backdoor GNN through model modification without data-poisoning}\\
\cmidrule(lr){1-3}
\multirow{3}{*}{\centering \parbox[c][6em][c]{1.8cm}{\centering Countermeasure Development}} 
  & \parbox[c][2em][c]{4.5cm}{\centering Black-Box Graph Backdoor Detection} 
  & \parbox[c][2em][c]{8.3cm}{$\diamond$ To detect GNN backdoor without available model info. or graphs} \\
\cline{2-3}
  & \parbox[c][2em][c]{4.5cm}{\centering Black-Box Graph Backdoor Mitigation} 
  & \parbox[c][2em][c]{8.3cm}{$\diamond$ To erase implanted backdoor without available model info. or graphs} \\
\cline{2-3}
  & \parbox[c][2em][c]{4.5cm}{\centering Graph Backdoor Explainability} 
  & \parbox[c][2em][c]{8.3cm}{$\diamond$ To develop explainability tools and algorithms for GNN backdoors} \\
\bottomrule
\end{tabular*}
\end{table*}

\subsection{Backdoor Optimization}
This kind of research primarily focuses on optimizing the stealthiness, and efficacy of GNN backdoors.
Several studies have transposed backdoor techniques from the realm of computer vision (CV) to graph learning. However, the subsequent sub-directions remain unexplored or warrant further enhancement.

\subsubsection{Graph Semantic Backdoor}
The semantic backdoor is a covert attack method where the attacker embeds specific semantic features (\textit{e.g.}, particular image or text patterns) in the training data to activate specific behaviors during inference. Unlike traditional backdoor attacks, semantic backdoors exploit the semantic information of the data rather than simple noise or obvious trigger patterns, which are harder to detect \cite{10.1145/3640333}.

Current GNN backdoor attacks also involve designing triggers with semantic information to infect training data, but the semantic information they carry is often adversary-specified features (\textit{e.g.}, Erdős-Rényi networks) or subgraph generated through heuristic methods (\textit{e.g.}, feature-optimized network generators), resulting in distribution differentiation between the trigger features and original clean data. This makes them easily identifiable by the defending side.

To optimize the incorporation of semantic backdoors into GNNs, several aspects could be considered:
\begin{itemize}
    \item 	Adaptive Trigger-optimization. 
    Future research could focus on adaptively integrating the statistical features of the target graph dataset and the parameters of the victim model to dynamically generate trigger subgraphs with internally-existing semantic information.
    Also, it may be feasible to introduce Generative Adversarial Networks (GANs) to generate trigger graphs with more realistic semantic information with provided model parameter and partial data.
    \item Hidden Semantic Features. 
    This aims to enhance the stealthiness of backdoors by embedding semantic information within the graph in a manner that keeps the infected graph data remains statistically indistinguishable from normal graph data. Consequently, this approach effectively reduces the likelihood of detection by defense mechanisms.
    
\end{itemize}

\subsubsection{Black-Box Graph Backdoor}
The black-box backdoor attack seeks to implant malicious backdoors into a model without accessing its training dataset, internal structure, or parameters. This type of attack leverages the understanding of the model's input-output relationship to introduce and exploit backdoors effectively \cite{9402020, 10.1145/3605212, 10.1145/3650205}.

There are two possible strategies for achieving black-box GNN attacks:

\begin{itemize}
    \item Inverse Engineering Approach. 
    To surmount the challenge of implanting backdoors without GNN information, adversaries could leverage black-box optimization algorithms (\textit{e.g.,} zeroth-order gradient descent or genetic algorithms) to generate alternative graphs and execute attacks.
    Additionally, to enhance the effectiveness of subgraph trigger design, various black-box graph explainability tools could be employed. 

    \item Surrogate Model Poisoning. 
    This idea utilizes the surrogate model of the victim GNN to generate infected data for executing attacks.
    Utilizing the black-box model stealing approach, the surrogate GNN could be constructed via model queries. Subsequently, conventional backdoor poisoning methods are employed to generate usable poisoned data, which is then injected into the training set of the victim model for data-poisoning attacks.

\end{itemize}

\subsubsection{Dormant Graph Backdoor}
Current graph backdoor attackers backdoor GNNs, and provide users with these compromised models directly. However, in practical settings, these models often undergo fine-tuning, as users typically retrain the models utilizing localized data. During such a process, the inherent backdoor may be easily corrupted \cite{10.1145/3319535.3354209}.

To optimize attacks, adversaries could develop and embed dormant backdoors in GNN models. Initially, these backdoors remain inactive and undetectable. However, they become operational subsequent to fine-tuning procedures (\textit{e.g.}, transfer learning).

To implement the dormant backdoor in GNN, one feasible way is to connect the trigger subgraph to an intermediate representation, which makes the representation produce a specific malicious output. 
To keep the backdoor inactive, the adversary could remove the related output layer information of that class before releasing the pre-trained model to the Internet. 
After fine-tuning for downstream tasks by victim users, the output layer replenishes the malicious-class-related information, thus resulting in the backdoor triggerable again.

\subsubsection{Untargeted Graph Backdoor}
In untargeted backdoor scenarios, adversaries aim to induce unpredictable effects on the model rather than targeting specific class. This type of backdoor can lead the model to produce incorrect predictions or exhibit unexpected behaviors without requiring the attacker to specify a precise target category or behavior \cite{10095980, 10323237}.

Combining untargeted backdoors in CV, to realize this in graph learning,
one straightforward approach is to embed graph triggers within the poisoned data, and carefully control the model's training process to maximize the feature distance between infected samples and their corresponding clean counterparts. This process forces, during testing, the attacked samples to be misclassified into non-targeted classes.

\subsection{Applicability Extension}
These studies aim to enhance the adaptability of GNN backdoor attacks across diverse application scenarios while ensuring the persistence of backdoor effectiveness.

\subsubsection{Generative Graph Backdoor}
Graph generative learning researches generating new graph structure by learning the latent structure and attribute distribution in the provided data, and the generative GNNs should enhance the diversity, rationality, and quality of generated graphs while preserving their structural and attribute information \cite{BONGINI2021242, pmlr-v139-lin21d}.

A prospective avenue for backdoor insertion in graph generative learning may implicate the implementation of generation condition-guided poisoning. 
Adversaries commence with the delineation of malicious generation conditions (trigger condition) and target behaviors. 
These triggers encompass a spectrum of graph features (\textit{e.g.}, node characteristics, edge connectivity patterns, or distinctive subgraph configurations).
Meanwhile, the design of target behaviors necessitates the definition of malicious features that the generated graph data will manifest when trigger conditions are met.
Examples of such features encompass the creation of molecules with deleterious properties, establishment of social network structures for the dissemination of misinformation, or fabrication of spurious information within a knowledge graph.
Subsequently, the adversary introduces trigger conditions and corresponding target behaviors into the training dataset, which facilitates the model's learning of the correlation between these triggers and the resultant malicious behaviors (\textit{i.e.}, get backdoored).

\subsubsection{Few-Shot Graph Backdoor}
In graph learning, utilizing few-shot learning serves the purpose of enhancing node classification, link prediction, and graph classification with a minimal number of graph structure samples. 
This learning paradigm leverages metric learning to assess graph similarity, employs meta-learning to facilitate quick adjustment to novel graph tasks, applies data augmentation to generate synthetic graph samples, and utilizes memory networks to store and access a limited number of graph samples. 
These approaches enhance the model's generalization capacity and adaptability when faced with data scarcity in graph learning scenarios \cite{garcia2018fewshot, Kim_2019_CVPR, Xie_2021_CVPR}. 

Combined with GNN, in few-shot graph learning, to implant the backdoor, initially, the adversary trigger could be selected based on specific combinations of node attributes, edge connection patterns, or unique subgraph structures to let them remain effective within a limited set of samples. 
Subsequently, synthetic graph samples containing the trigger conditions are generated using data augmentation to expand the few-shot training data. 
And then, models are trained using meta-learning (\textit{e.g.}, MAML or metric learning like Prototypical Networks) to enable rapid adaptation and learning of the relationship between trigger conditions and malicious behaviors with only a small number of samples. 
Throughout the training, the GNN model gets backdoored and presets malicious behaviors if the input satisfies the backdoor activation conditions (\textit{e.g.}, input poisonous data with trigger subgraphs). 

\subsubsection{Graph Large Language Model Backdoor}
The Graph Large Language Model (GLLM) synergizes the capabilities of GNN and large language model through joint training and multimodal learning to comprehensively handle graph-structured data processing and natural language generation. 
Technically, GLLM integrates graph and text embeddings utilizing shared representation layers, attention mechanisms, etc. 
By optimizing model performance through fine-tuning, distillation, and compression, GLLM is widely deployed in various domains such as knowledge graphs, recommendation systems, economic analysis, drug discovery, etc. \cite{10.1145/3655103.3655110, Besta_Blach, Tian_Song}.

There are several possible research directions for implanting backdoors or its malicious application in GLLMs:
\begin{itemize}
    \item Fine-tuning Backdoor Embedding. 
    Due to the substantial training costs, fine-tuning has become the predominant method for training large models, which could be generally divided into two main kinds: Full Fine-tuning and Repurposing. For
    considerations of time saving, most users and MLaaS companies prefer Repurposing, which comprises keeping the majority of the model parameters fixed and only updating parameters relevant to the output \cite{10470654, Zhang_Huang_Liu_Tan}. 
    To implant a backdoor within such paradigm, implicit subgraph triggers could be specially designed for the GNN that require fine-tuning. 
    Through injecting these triggers into fine-tuning data and proceeding with mode-tuning, the finally updated parameters could cause the trained model to exhibit specific behaviors when encountering malicious triggers in test. 
    To enhance attack, valuable research points implicate efficient poisoning of large-scale datasets, backdoor injection through multi-task fine-tuning, explainability of backdoors in large models. 
    
    \item Distributed Backdoor Embedding. 
    Distributed fine-tuning maximizes computing resources across multiple devices, handling large datasets, and adapting to various hardware platforms and network conditions to meet the demands of large-scale deployment \cite{10476280, 10446717}.
    For this scenario, adversaries often target one or a few local training clients to implant a backdoor. 
    Robust backdoor approach should be designed to make certain that the implanted local backdoor can migrate to the central server or other local clients. Moreover, adversaries should keep the stealthiness of the attack, and infected data should not differ significantly in distribution from other local data where data distribution is often non-IID (thus avoiding anomaly detection and filtration).

    \item Data Augmentation Backdoor Embedding. 
    Fine-tuning data augmentation is a technique of transforming or expanding training data during the fine-tuning stage, which enhances data diversity and improves model generalization \cite{osti_10448809, moller-etal-2024-parrot}.
    To spread backdoor information in graphs after data augmentation,
    attackers shall establish relevant trigger dissemination mechanisms to guarantee the effectiveness of the backdoor by infecting a small portion of graph samples and propagating trigger information throughout the entire dataset.
    


    \item Multi-modal Graph Backdoor Embedding. 
    It refers to the implementation of backdoor attacks by exploiting multiple data types including graphs, images, and text simultaneously within the multi-modal large model. To implant backingdoor through data-poisoning, several points merit attention: research on effectively integrating features from different modalities to simultaneously represent poisonous graphs and text data; investigating how to transfer information between different modalities to achieve more effective attacks during the backdoor injection process; designing the backdoor mechanism that is triggered by multi-modal data, allowing it to simultaneously use information from different modalities for activation.

\end{itemize}

\subsubsection{Parameter Modification Backdoor}
Most GNN backdoor attacks are carried out through poisoning, aiming to manipulate the model to produce specific outputs in response to certain trigger features specified by the attacker. 
To achieve this goal, adversaries may directly modify GNN parameters to achieve this effect \cite{NEURIPS2022_3538a22c}.

To achieve this, it is necessary to utilize 
explainability tools to analyze the GNN parameters and understand their sensitivity to various features. 
Consequently, corresponding parameter modification can be implemented to make the model responsive to generate adversary-desired results when given designated subgraph trigger modes.

\subsection{Countermeasure Development}
The principal objective of backdoor defenses is to prevent the activation of backdoors or to completely eliminate them from the model.

There are three primary defense strategies against backdoors in deep learning: (\textit{i}) backdoor detection; (\textit{ii}) sample filtering; (\textit{iii}) model backdoor mitigation. 
The first type aims to ascertain whether the suspicious model has been backdoored.
The second strategy entails the filtration of malicious parts in input samples to neutralize triggers and inhibit backdoor behaviors.
The third category targets eliminating embedded backdoors within models by leveraging existing samples and backdoored models.

\subsubsection{Black-Box Graph Backdoor Detection}
Current backdoor detection methodologies require access to training graphs and model parameters. However, to safeguard user privacy, defenders are typically unable to obtain such data.
Consequently, the development of black-box detection methodology that do not rely on available samples and model parameters has become essential \cite{Dong_2021_ICCV, guo2022aeva}.

Drawing on existing defenses in CV, the black-box detection of backdoors in GNN could be accomplished through reverse engineering for usable poisonous graphs or potential triggers. 
This procedure could effectively be performed through non-gradient-based approaches (\textit{e.g.}, graph ant algorithm or query-based optimization), and following this, extreme value analysis might be utilized to verify the accuracy of the reversed triggers, which enables further identification of the target class of the attack and the extraction of other relevant information (\textit{e.g.}, parameter changes or impact range).

\subsubsection{Black-Box Graph Backdoor Mitigation}
Based on the results of backdoor detection, users frequently petition the defense side for the removal of embedded backdoors within the model in instances where data and model parameters are unavailable (\textit{i.e.}, black-box conditions) for privacy protection \cite{Wang_Ya_2024, leroux:hal-04485197}.

To achieve this, two potential ideas may be considered: 
the first one encompasses employing black-box reverse engineering to synthesize artificial graphs, followed by retraining or fine-tuning the GNN using these data alongside a constrained loss; 
The second method utilizes the concept of data distillation, wherein a new clean model is generated from the suspicious model. This new model aims to inherit normal task accuracy from the backdoored GNN while eliminating its backdoor characteristics.

\subsubsection{Graph Backdoor Explainability}
Compared to backdoor research in the field of CV, research on GNN backdoors is still in its nascent stage.
Moreover, due to the non-Euclidean structural nature of graph data, the definition and explainability of GNN backdoors have yet to be fully elucidated. 
Therefore, to aid defenders in understanding the essence of backdoors and to foster related research, explainability tools and visualization platforms shall be further developed, along with more in-depth analyses of various backdoor characteristics.


\section{Conclusion}
Graph neural backdoors have emerged as a nascent but rapidly evolving field since 2021. 
Despite its promising developments, a comprehensive investigation of this area remains critically lacking.
To bridge this gap, in this survey, various aspects of graph neural backdoors are thoroughly examined, including the technical foundations, existing attack mechanisms, and corresponding defense strategies.
Furthermore, we explore potential benign applications of this backdoor technology and possible future research directions. 
We aim to provide valuable and timely insights for defenders to better address future threats, inspire more researchers to focus on neural backdoor issues, and promote further safety development in this field.
With the advancement of graph learning and the proliferation of MLaaS in AI industry, it is imperative to advocate for secure and robust graph learning research.

\bibliographystyle{ieeetr}
\bibliography{ref}

\vfill

\end{document}